\documentclass{article}

    \usepackage[nonatbib,final]{neurips_2020}

\usepackage[utf8]{inputenc} 
\usepackage[T1]{fontenc}    
\usepackage{hyperref}       
\usepackage{url}            
\usepackage{booktabs}       
\usepackage{amsfonts}       
\usepackage{nicefrac}       
\usepackage{microtype}      

\usepackage{graphicx}
\usepackage{xcolor}
\usepackage{todonotes}
\usepackage{multirow} 
\usepackage{caption}
\usepackage{subfigure}
\usepackage{float}
\usepackage{xspace}

\title{ForestNet: Classifying Drivers of Deforestation in Indonesia using Deep Learning on Satellite Imagery}

\author{%
    Jeremy Irvin\thanks{Equal contribution.} ~$^1$, Hao Sheng\footnotemark[1]~~$^1$, Neel Ramachandran$^1$, Sonja Johnson-Yu$^1$\\ 
    \textbf{Sharon Zhou$^1$, Kyle Story$^2$, Rose Rustowicz$^2$,  Cooper Elsworth$^2$} \\ 
    \textbf{Kemen Austin\thanks{Equal contribution.} ~$^3$, Andrew Y. Ng\footnotemark[2]~~$^1$ }\\
    $^1$Stanford University, $^2$Descartes Labs, $^3$RTI International\\
    \small{\texttt{\{jirvin16,haosheng,neelr,sonjyu,sharonz\}}\texttt{@cs.stanford.edu}} \\
    \small{\texttt{\{kyle,rose,cooper\}}\texttt{@descarteslabs.com}},
    \small{\texttt{kaustin@rti.org}},   \small{\texttt{ang@cs.stanford.edu}}
}

\begin{document}

\maketitle

\newif\ifshowcomments
\showcommentsfalse
\ifshowcomments
\newcommand\jeremy[1]{\textcolor{red}{[jirvin: #1]}}
\newcommand\hao[1]{\textcolor{blue}{[hao: #1]}}
\newcommand\neel[1]{\textcolor{green}{[neel: #1]}}
\newcommand\sonja[1]{\textcolor{purple}{[sonja: #1]}}
\else
\newcommand\jeremy[1]{}
\newcommand\hao[1]{}
\newcommand\neel[1]{}
\newcommand\sonja[1]{}
\fi

\newcommand{\modelname}{ForestNet\xspace}

 \begin{abstract}
Characterizing the processes leading to deforestation is critical to the development and implementation of targeted forest conservation and management policies. In this work, we develop a deep learning model called \modelname to classify the drivers of primary forest loss in Indonesia, a country with one of the highest deforestation rates in the world. Using satellite imagery, \modelname identifies the direct drivers of deforestation in forest loss patches of any size. We curate a dataset of Landsat 8 satellite images of known forest loss events paired with driver annotations from expert interpreters. We use the dataset to train and validate the models and demonstrate that \modelname substantially outperforms other standard driver classification approaches. In order to support future research on automated approaches to deforestation driver classification, the dataset curated in this study is publicly available at \url{https://stanfordmlgroup.github.io/projects/forestnet}.
 \end{abstract}

\section{Introduction}
The preservation of forests is crucial for preventing loss of biodiversity, managing air and water quality, and mitigating climate change \cite{foley2005global,foley2011solutions}. Forest loss in the tropics, which contributes to around 10\% of annual global greenhouse gas emissions \cite{arneth2019ipcc}, must be reduced to decrease the potential for climate tipping points \cite{lenton2019climate}. The direct drivers of tropical deforestation, or the specific activities which lead to forest cover loss, include natural events like wildfires, as well as human-induced land uses such as industrial and smallholder agricultural development \cite{hosonuma2012assessment}. Determining the extent to which these processes contribute to forest loss enables companies to fulfill their zero-deforestation commitments and helps decision-makers design, implement, and enforce targeted conservation and management policies \cite{donofrio2017supply,henders2018national,seymour2019reducing,hansen2020policy}. Indonesia has one of the highest rates of primary forest loss in the tropics that places it among the largest emitters of greenhouse gases worldwide \cite{austin2018review}. Recent work using high-resolution satellite imagery to manually classify the direct drivers of forest loss in Indonesia provides critical information about the nationwide causes of deforestation \cite{austin2019causes}. Methods to automate driver classification would enable more spatially-broad and temporally-dense driver attribution with significant implications on forest conservation policies.

The growth in availability of high-resolution satellite imagery coupled with advancements in deep learning methods present opportunities to automatically derive geospatial insights at scale \cite{karpatne2018machine,janowicz2020geoai}. Recently, convolutional neural networks (CNNs) have enabled a variety of satellite-image based applications, including extracting socioeconomic data for tracking poverty \cite{jean2016combining}, classifying land use for urban planning \cite{albert2017using}, and identifying the locations of solar photovoltaics in the U.S. to support the adoption of solar electricity \cite{yu2018deepsolar}. Prior work has developed decision-tree-based models and other machine learning techniques to automatically identify land-use conversion following deforestation \cite{richards2016rates,curtis2018classifying,phiri2019long,descals2019oil,poortinga2019mapping,hethcoat2019machine}. However, these methods do not leverage the complete contextual information provided by satellite imagery and heavily depend on data that is not widely available at high resolution. 

\begin{table*}[t!]
    \centering
    \resizebox{0.5\textwidth}{!}{%
    \begin{tabular}{l|c|c|c}
      \toprule
      Driver Class, N & Training & Validation & Test\\
      \midrule
      Plantation & 686 & 219 & 265\\
       Smallholder Agriculture & 556 & 138 & 207\\
      Grassland/shrubland & 143 & 47 & 85\\
      Other & 231 & 70 & 112\\
      \midrule
      Overall & 1,616 & 474 & 669\\
      \bottomrule
    \end{tabular}
  }
  \caption{Number of events per driver class in the training, validation, and test sets. Each class was created by combining driver categories from \cite{austin2019causes} as shown in Table~\ref{table:grouping} in the Appendix.}
  \label{table:stats}
\end{table*}

In this work, we train CNNs to identify the direct drivers (plantation, smallholder agriculture, grassland/shrubland, or other drivers) of primary forest loss using satellite imagery in Indonesia. This task is challenging to automate due to the heterogeneity of drivers within images and driver classes, the rapid evolution of landscapes over time, and the lack of expert-annotated data. We highlight three key technical developments that enable high classification performance on this task, including pre-training on a large land cover dataset that we curate, data augmentation using satellite revisits, and multi-modal fusion with auxiliary predictors to complement the satellite imagery. Our best model, called \modelname, substantially outperforms standard driver classification approaches. All data used in this study is publicly available to support future research on forest loss driver classification.

\section{Methods}

\subsection{Forest Loss Events and Driver Annotations}
The coordinates of forest loss events and driver annotations used in this work were curated in \cite{austin2019causes}. Global Forest Change (GFC) published maps were used to obtain a random sample of primary natural forest loss events at 30m resolution from 2001 to 2016 \cite{hansen2013high}. Each forest loss region is represented as a polygon and is associated with a year indicating when the forest loss event occurred. An expert interpreter annotated each event with the direct driver of deforestation using high-resolution satellite imagery from Google Earth \cite{austin2019causes}. 

We group the drivers into categories that are feasible to differentiate using 15m resolution Landsat 8 imagery and to ensure sufficient representation of each category in the dataset. Due to the availability of Landsat 8 imagery starting in 2013 and challenges with constructing clear images with Landsat 7, we additionally remove any loss events before 2012 for driver classes that are likely to change over time \cite{austin2019causes}. Driver groupings and classes for which forest loss events before 2012 were dropped are shown in Table~\ref{table:grouping}, and examples of each class are provided in Figure~\ref{fig:examples} in the Appendix.

The complete dataset consists of 2,756 forest loss events with driver annotations. The dataset was randomly split into a training set to learn model parameters, a validation set to compare models, and a test set to evaluate the best model (Table~\ref{table:stats}, Figure~\ref{fig:splits_drivers}). All examples in the validation and test sets were manually reviewed, and labels were corrected if the original label did not accurately represent the forest loss event by examination in high-resolution imagery using Google Earth.

\subsection{Satellite Imagery}
We capture each forest loss region with Tier 1 Landsat 8 satellite imagery acquired within five years of the event's occurrence. All images are centered around the forest loss regions with visible bands pan-sharpened to 15m per-pixel resolution and a total size of 332 x 332 pixels, corresponding to an area of 5 square kilometers (500 hectares). Images for forest loss events occurring between 2012 and 2016 are captured during the four-year period starting the year after the event occurred (e.g., 2015-2018 for a 2014 event). Images for forest loss events occurring prior to 2012 are captured between 2013 and 2016 due to the availability of Landsat 8 imagery beginning in 2013. All images are converted to surface reflectance to account for atmospheric scattering or absorption.

We search for individual cloud-filtered scenes that capture the region of interest within the time range. We minimize cloud cover by only considering images with less than 50\% cloudy pixels and 0\% cirrus pixels according to the native Landsat 8 cloud and cirrus bands, respectively. We also construct a composite image by taking a per-pixel median over these cloud-filtered scenes, using the five least cloudy scenes when less than five such scenes were available. Using this procedure, we obtain exactly one composite image for each example and additional images for any individual cloud-filtered scenes.

\subsection{Baseline Models and Auxiliary Predictors}
Following recent work on automatically classifying land-use conversion from deforestation \cite{richards2016rates,phiri2019long},
we develop random forest (RF) models that input a variety of variables including topographic, climatic, soil, accessibility, proximity, and spectral imaging predictors (Table~\ref{table:predictors}). We additionally develop models which only input the Landsat 8 visible bands to assess the benefit of including the auxiliary variables. Details of the baseline models are provided in the Appendix. 

\subsection{\modelname}

\textbf{Task.}
The deforestation driver classification task is to identify the direct driver of deforestation in a region of primary forest loss. Instead of a canonical multi-class classification approach, we formulate the task as semantic segmentation to (1) address that there are often multiple land uses within a single image, (2) implicitly utilize the information specific to the loss region, and (3) allow for high resolution (15m) predictions that can be used to predict different drivers for multiple loss regions of varying sizes. At train-time, we assign the single driver label to all of the pixels within the forest loss region. At test-time, the per-pixel model predictions in the forest loss region are used to obtain a single classification of the whole region, described below.

\textbf{Architecture and Pre-Training.}
We experiment with a variety of segmentation architectures, including UNet \cite{ronneberger2015u}, Feature Pyramid Networks \cite{lin2017feature}, and DeepLabV3 \cite{chen2017rethinking}. For each segmentation architecture, we try several backbone architectures, including variants of ResNet \cite{he2016deep}, DenseNet \cite{huang2017densely}, and EfficientNet \cite{tan2019efficientnet}. We experiment with random weight initialization and pre-trained initialization learned from a large land cover dataset in Indonesia. We denote the use of pre-training as PT, and details of the training procedure, pre-training procedure, and loss functions are in the Appendix.



\begin{figure*}[t!]
  \centering
    \includegraphics[scale=0.22]{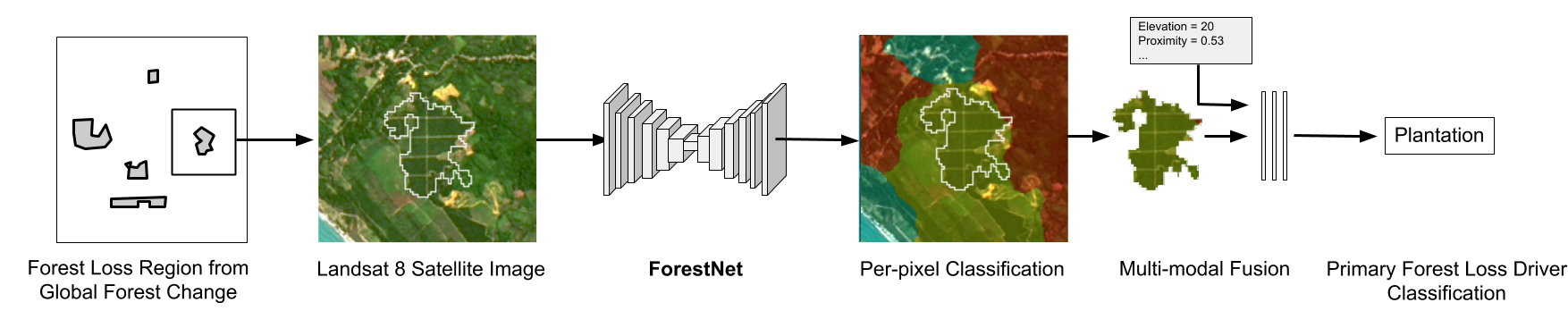}
  \caption{
     ForestNet inputs a Landsat 8 satellite image centered around a region of forest loss and identifies the direct driver of forest loss in that region.
  }
  \label{fig:model}
\end{figure*}

\textbf{Data Augmentation.}
We experiment with several data augmentation techniques, including affine transformations, salt and pepper noise, and artificial clouds, haze, and fog \cite{imgaug}. We additionally randomly sample from the scenes and composite images during training to capture changes in landscape, which can occur in the same location due to seasonal differences, for example. We refer to this procedure as \textit{scene data augmentation} (SDA). Images are randomly cropped to 160 x 160 pixels (approximately 2.4 x 2.4 km) during training, and center cropped to that size during prediction. No loss is computed for examples where the loss region is fully truncated due to cropping.

\textbf{Multi-Modal Fusion with Auxiliary Variables.} 
To leverage the auxiliary predictors in the deep learning models, we explore the use of multi-modal fusion \cite{atrey2010multimodal, Sheng_2020_CVPR_Workshops}. Specifically, we compute the mean, standard deviation, minimum, and maximum of the predictors in Table~\ref{table:predictors} within the forest loss region, concatenate them to the segmentation logits, then feed the result through three fully connected layers to obtain the class logits. We include ReLU and dropout after the first and second layers.

\textbf{Driver Classification.}
The image captured temporally closest to the year of the forest loss event is input to the model at test-time to produce a per-pixel classification of the direct drivers within the forest loss region. The composite is used when no single scene meets the quality requirements. In order to convert the per-pixel logits within the forest loss region into a single classified driver, we compute the mean of the per-pixel scores in the polygon and make the predicted class the one assigned the highest score in the mean. For multi-modal fusion, the predicted class is the one assigned the highest score in the output of the final fully connected layer.

\section{Results}

\textbf{Performance Measures on the Validation Set.} The RF model that only inputs data from the visible Landsat 8 bands achieved the lowest performance on the validation set, but the incorporation of auxiliary predictors substantially improved its performance (Table~\ref{table:test_results} left). All of the CNN models outperformed the RF models. The best performing model, which we call \modelname, used an FPN architecture with an EfficientNet-B2 backbone. The use of SDA provided large performance gains on the validation set, and land cover pre-training and incorporating auxiliary predictors each led to additional performance improvements. 

\captionsetup[table]{skip=3pt}
\begin{table}[t!]
  \centering
  \resizebox{0.95\textwidth}{!}{%
    \begin{tabular}{l|c|cc|cc}
      \toprule
      \multirow{2}{*}{Model} & \multirow{2}{*}{Predictors} & \multicolumn{2}{c|}{Val} & \multicolumn{2}{c}{Test}\\
      & & Acc & F1 & Acc & F1\\
      \midrule
      RF & Visible & 0.56 & 0.49 & 0.49 & 0.44\\
      RF & Visible + Aux & 0.72 & 0.67 & 0.67 & 0.62 \\
      CNN & Visible & 0.80 & 0.75 & 0.78 & 0.70 \\
      CNN + SDA & Visible & 0.82 & 0.79 & 0.78 & 0.73\\
      CNN + SDA + PT & Visible & 0.83 & 0.80 & 0.80 & 0.74\\
      CNN + SDA + PT & Visible + Aux & {\bf 0.84} & {\bf 0.81} & {\bf 0.80} & {\bf 0.75} \\
      \bottomrule
    \end{tabular}
    \quad
    \begin{tabular}{l|c|c|c}
      \toprule
      Driver Category & P & R & F1\\
      \midrule
      Plantation & 0.81 & 0.92 & 0.86\\
      Smallholder Agriculture & 0.81 & 0.77 & 0.79\\
      Grassland/shrubland & 0.59 & 0.57 & 0.58\\
      Other & 0.90 & 0.70 & 0.79\\
      \midrule
      Macro-average & 0.78 & 0.74 & 0.75\\
      \bottomrule
    \end{tabular}
  }
  \caption{Accuracy and macro-average of the per-class F1 scores of the baseline and model variants on the validation set (left) and per-class performance metrics of ForestNet on the test set (right), including precision (P), recall (R), and F1-score. Validation results are reported as the average of 10 runs, and test results are reported for the best validation run.}
  \label{table:test_results}
\end{table}

\textbf{Performance Measures on the Test Set.}
\modelname achieved high classification performance across the four-driver categories on the test set (Table~\ref{table:test_results} right). The model achieved the best performance metrics on plantation and smallholder agriculture, which were the classes with the highest prevalence in the dataset. \modelname performance on grassland/shrubland on the test set was the lowest primarily due to confusion with smallholder agriculture examples.

\section{Discussion}
We developed a deep learning model called \modelname to automatically identify the direct drivers of primary forest loss in satellite images. To our knowledge, this is the first study to use deep learning for classifying the drivers of deforestation. A recent study has developed a semantic segmentation model for mapping industrial smallholder plantations, but is limited to closed-canopy oil palm and does not focus on land use directly after deforestation \cite{descals2020high}.

This study has important limitations that should be considered. First, \modelname only leverages a single satellite image to make the classification, but the evolution of a landscape over time is important for identifying the direct driver \cite{austin2019causes}. Second, the model cannot differentiate between different species of plantations and types of smallholder agriculture development. Future work should explore the use of multiple high-resolution images to improve the accuracy and granularity of the driver predictions.

Our work contributes to the growing effort to use machine learning for tackling problems relevant to climate change \cite{rolnick2019tackling}. \modelname has the potential to generate accurate, temporally, and spatially dense maps of forest loss drivers over the entire nation of Indonesia. This new data could aid policymakers in developing more effective forest conservation and management policies to combat deforestation, one of the major contributors to global greenhouse gas emissions \cite{arneth2019ipcc}. Focusing on forest loss in Indonesia is particularly important for climate change as the role of deforestation in its emissions profile is significantly larger than the rest of the world \cite{republic2017indonesia}. We hope that the methodology and data presented in this work eventually lead to comprehensive maps of forest loss drivers worldwide.


\small
\bibliography{bibliography}
\bibliographystyle{ieeetr}

\newpage
\appendix
\section*{Appendix}
\subsection*{Dataset}
\setcounter{table}{0}
\setcounter{figure}{0}
\renewcommand{\thetable}{A\arabic{table}}
\renewcommand{\thefigure}{A\arabic{figure}}
\begin{table}[ht]
  \begin{center}
    \begin{tabular}{l|l|c}
      \toprule
      Driver Group & Driver Category & Drop Pre-2012\\
      \midrule
      \multirow{3}{*}{Plantation} & Oil palm plantation & No\\
      & Timber plantation & No\\
      & Other large-scale plantations & No\\
      \midrule
      Grassland/shrubland & Grassland/shrubland & Yes\\
       \midrule
      \multirow{3}{*}{Smallholder Agriculture} & Small-scale agriculture & No\\
      & Small-scale mixed plantation & No\\
      & Small-scale oil palm plantation & No\\
       \midrule
      \multirow{5}{*}{Other} & Mining & No\\
      & Fish pond & No\\
      & Logging road & Yes\\
      & Secondary forest & Yes\\
      & Other & No\\
      \bottomrule
    \end{tabular}
  \end{center}
  \caption{The driver categories defined in \cite{austin2019causes} are merged into groups that are feasible to differentiate using 15m resolution Landsat 8 imagery and to ensure sufficient representation of each category in the dataset. We additionally remove any loss events before 2012 for driver classes that are likely to change over time \cite{austin2019causes}.}
  \label{table:grouping}
\end{table}
\begin{figure*}[ht!]
  \centering
  \includegraphics[width=\linewidth]{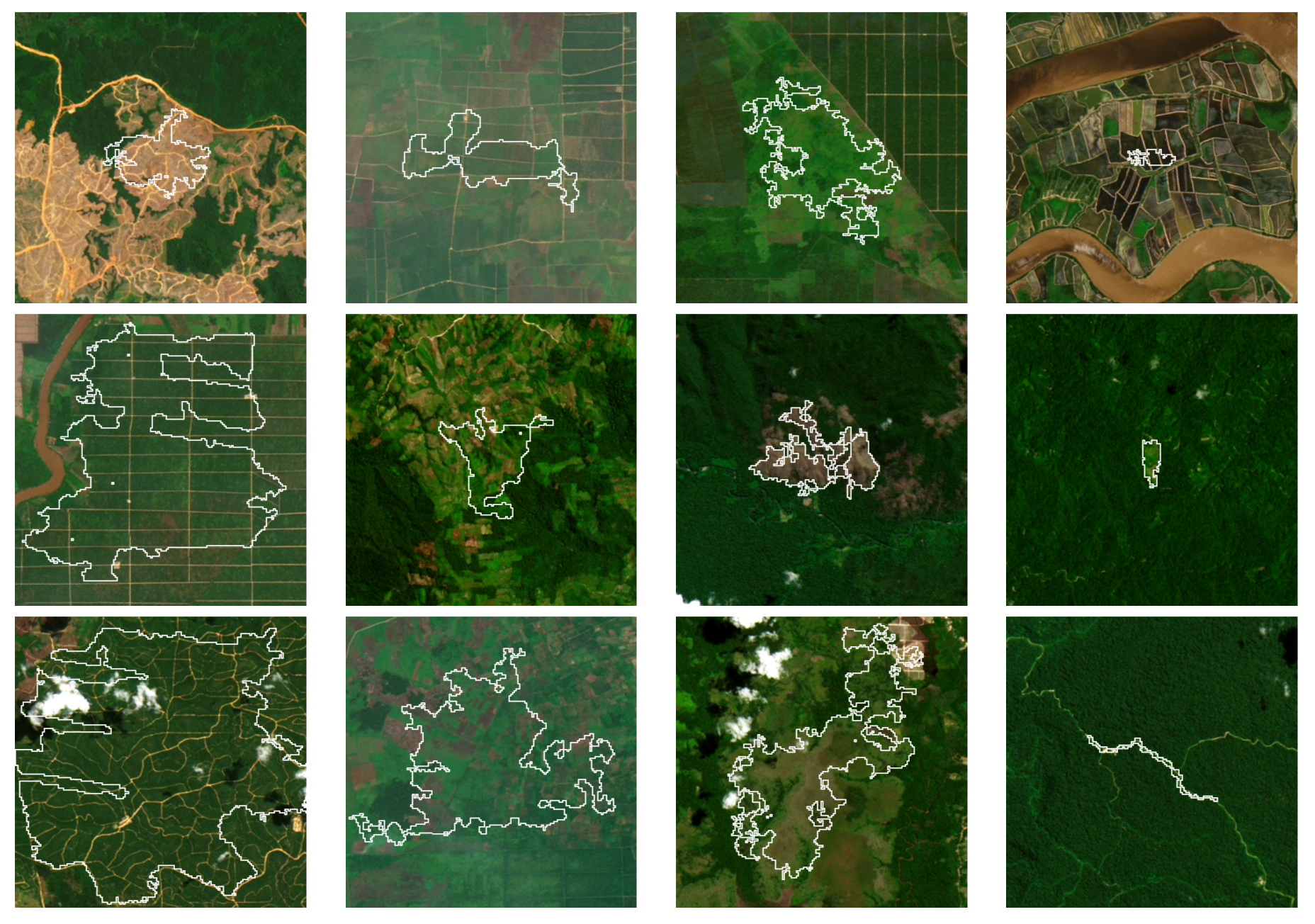}
  \caption{Characteristic examples of the four driver classes: Plantation (far left column), Smallholder agriculture (middle left column), Grassland/shrubland (middle right column), Other (far right column). Forest loss regions overlay the images here for visualization, but do not overlay the images input to the model.
  }
  \label{fig:examples}
\end{figure*}
\setlength{\tabcolsep}{0.25em}
\begin{table}[ht]
  \begin{center}
   \resizebox{\textwidth}{!}{
    \begin{tabular}{c|c|c|c|c|c}
      \toprule
      \multirow{2}{*}{Predictor Group} & \multirow{2}{*}{Predictor (units)} & Spatial & Temporal & \multirow{2}{*}{Source} & \multirow{2}{*}{References}\\
      & & Resolution & Resolution & & \\
      \midrule
      \multirow{3}{*}{Topographic} & Elevation (m) & \multirow{3}{*}{30m} & \multirow{3}{*}{N/A} & \multirow{3}{*}{USGS (SRTM)} & \multirow{3}{*}{\cite{srtm}}\\
       & Slope (0.01$^{\circ}$) & & & &\\
       & Aspect (0.01$^{\circ}$) & & & &\\
       \midrule

       
      \multirow{16}{*}{Climatic} & Surface-Level Albedo (0.01\%)& \multirow{16}{*}{56km} &  \multirow{16}{*}{1 day}& \multirow{16}{*}{NCEP (CFSv2)} & \multirow{16}{*}{\cite{Saha2014}}\\
      &  Clear-Sky Longwave Flux (W/m$^2$)&  &  & &\\
      &  Clear-Sky Solar Flux (W/m$^2$)& & & &\\
      & Direct Evaporation from Bare Soil (W/m$^2$) & & & &\\
      & Longwave Radiation Flux (W/m$^2$)& & & &\\
      & Shortwave Radiation Flux (W/m$^2$) & & & &\\
      & Ground Heat Net Flux (W/m$^2$) & & & &\\
      & Latent Heat Net Flux (W/m$^2$) & & & &\\
      & Specific Humidity ($10^{-4}$ kg/kg) & & & &\\
      & Potential Evaporation Rate (W/m$^2$) & & & &\\
      & Ground-Level Precipitation (0.1 mm) & & & &\\
      & Sensible Heat Net Flux (W/m$^2$) & & & &\\
      & Volumetric Soil Moisture Content (0.01\%) & & & &\\
      & Air Pressure at Surface Level (10 Pa) & & & &\\
      & Wind Components 10m above Ground (0.01 m/s) & & & &\\
      & Water Runoff at Surface Level (0.01 kg/m$^2$) & & & &\\
       
       \midrule
      Soil & Presence of Peat & N/A & N/A & GFW (MoA) & \cite{peat}\\
      \midrule
      Accessibility & Euclidean Distance to Road (km) & N/A & N/A & Open Street Map & \cite{OpenStreetMap}\\
      \midrule
      Proximity &  Euclidean Distance to City (km) & N/A & N/A & Open Street Map & \cite{OpenStreetMap}\\
      \midrule
      \multirow{3}{*}{Imaging} & Landsat 8 Visible & 15m & \multirow{3}{*}{16 days} & \multirow{3}{*}{USGS (Landsat 8)}  & \multirow{3}{*}{\cite{Landsat8}} \\
      & Landsat 8 IR & 30m & & \\
      & Landsat 8 NDVI & 30m & & \\
      \bottomrule
    \end{tabular}
  }
  \end{center}
  \caption{The baseline models incorporate a variety of predictors that are commonly used in automatic methods for classifying land use. The climatic predictors were aggregated over the five years before the event using a mean, minimum, and maximum over the daily values. The imaging predictors were aggregated using the procedure to construct images with minimal cloud cover. All other predictors were measured at a single point in time.}
  \label{table:predictors}
\end{table}
\begin{figure*}[ht!]
  \centering
    \includegraphics[width=0.8\linewidth]{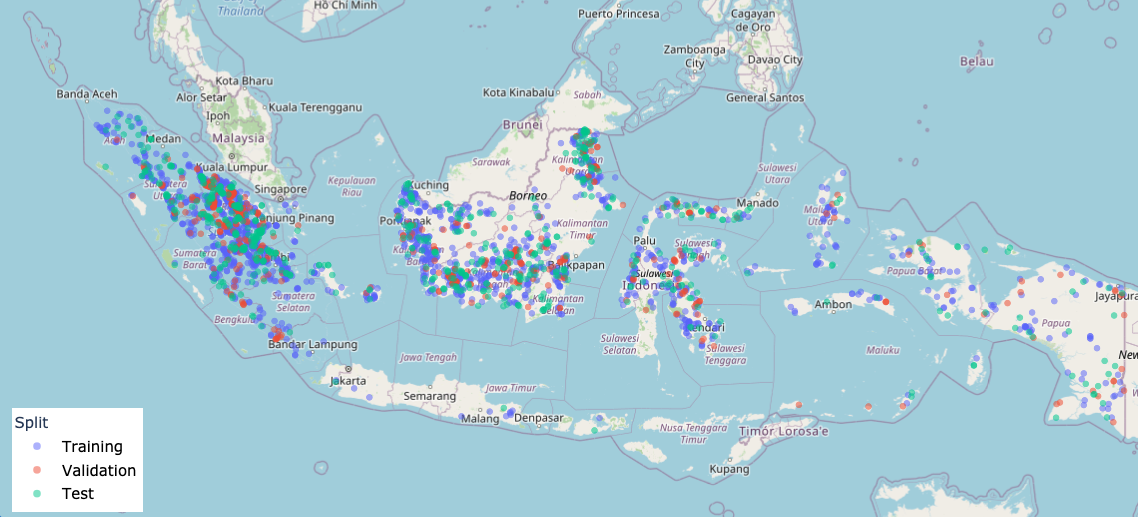} \\
    \includegraphics[width=0.8\linewidth]{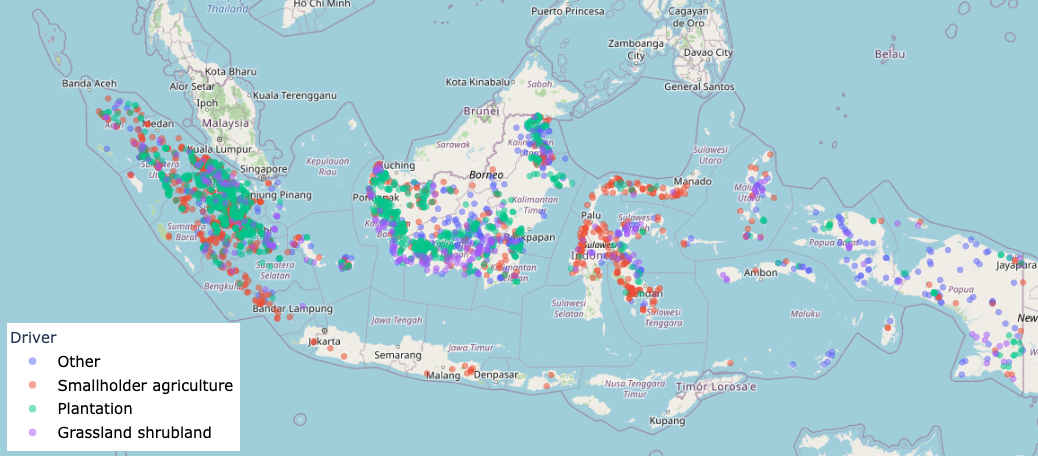}
  \caption{Spatial distribution of training, validation, and test splits across Indonesia (top) and spatial distribution of driver class across Indonesia (bottom). The training, validation, and test sets were sampled to ensure no spatial overlap between images in different splits.}
  \label{fig:splits_drivers}
\end{figure*}

\subsection*{Baseline Model Details}
\captionsetup[table]{skip=3pt}
\begin{table}[ht]
  \centering
  \resizebox{0.62\textwidth}{!}{%
    \begin{tabular}{l|c|cc|cc}
      \toprule
      \multirow{2}{*}{Model} & \multirow{2}{*}{Predictors} & \multicolumn{2}{c|}{Val} & \multicolumn{2}{c}{Test}\\
      & & Acc & F1 & Acc & F1\\
      \midrule
      \multirow{2}{*}{Logistic Regression Classifier} & Visible & 0.58 & 0.39 & 0.52 & 0.39\\
       & Visible + Aux & 0.70 & 0.61 & 0.60 & 0.52 \\
      \midrule
      \multirow{2}{*}{Ridge Regression Classifier} & Visible & 0.59 & 0.38 & 0.50 & 0.33\\
       & Visible + Aux & 0.70 & 0.58 & 0.62 & 0.51 \\
      \midrule
      \multirow{2}{*}{K-nearest Neighbor} & Visible & 0.53 & 0.39 & 0.46 & 0.33\\
       & Visible + Aux & 0.61 & 0.45 & 0.56 & 0.47 \\
      \midrule
       \multirow{2}{*}{Multi-layer Perceptron} & Visible & 0.51 & 0.36 & 0.52 & 0.40\\
       & Visible + Aux & 0.68 & 0.55 & 0.58 & 0.50 \\
       \midrule
        \multirow{2}{*}{Decision Tree} & Visible & 0.40 & 0.38 & 0.37 & 0.37\\
       & Visible + Aux & 0.58 & 0.53 & 0.52 & 0.46 \\
      \midrule
      \multirow{2}{*}{Random Forest} & Visible & 0.56 & 0.49 & 0.49 & 0.44\\
       & Visible + Aux & \textbf{0.72} & \textbf{0.67} & \textbf{0.67} & \textbf{0.62} \\
      \bottomrule
    \end{tabular}
  }
  \caption{Accuracy and macro-average of the per-class F1 scores of all baseline models on the validation and test sets.}
  \label{table:extra_baseline_results}
\end{table}
We train multiple baseline models, including a decision tree, random forest, logistic regression classifier, ridge regression classifier, k-nearest neighbor classifier, and a multi-layer perceptron, and perform hyperparameter tuning using 3-fold cross-validation. For the tree-based models (decision tree and random forest), we tune the max depth of each tree, the minimum samples in leaves, and the number of decision trees. For the linear models (logistic regression classifier and ridge regression classifier), we tune the strength of regularization and the regularization norm (L1 or L2). For the k-nearest neighbor classifier, we tune the number of nearest neighbors. For the multi-layer perceptron model, we tune the number of hidden layers, the number of neurons in each hidden layer, and the learning rate.

We train two types of baseline models. The first type inputs and outputs data at 15m resolution, and all pixels within the loss regions in the training set were used as separate examples to train the model. Once the model was fit, a single driver classification over the full loss region is obtained from the model using the mode prediction over all pixels in the region. The second type of model inputs and outputs data corresponds to the forest loss region, where the per-pixel inputs are aggregated using a variety of statistics, including the mean, standard deviation, minimum, and maximum value of the variables within the region. The region-based model outperformed the pixel-based model, so we report the region-based model performance in Table~\ref{table:test_results} and Table~\ref{table:extra_baseline_results}.

\subsection*{CNN Loss Function}
We experiment with a variety of segmentation losses, including cross entropy loss, focal loss \cite{lin2017focal}, generalized dice loss \cite{sudre2017generalised}, and a convex combination of focal and dice loss. Due to the heterogeneity of land use within a single image, we only compute segmentation loss on pixels within the forest loss region to avoid incorrect supervision on many pixels. We additionally include a cross-entropy loss that takes the mean of the forest loss region logits as input for the non-multi-modal models, and the output of the final fully connected layer for the multi-modal models. The final loss is a linear combination of the segmentation and classification losses.

\subsection*{CNN Training Procedure} 
For the CNN models, extensive hyperparameter tuning was performed over learning rate, architecture, backbone, data augmentation, regularization (dropout and weight decay), focal loss focusing parameter, coefficient of the convex combination in the segmentation loss, coefficient of the cross-entropy loss in the linear combination of the segmentation and classification loss, and batch size. During training, we evaluate the model on the validation set after each epoch and save the checkpoint with the highest F1 score averaged over the four-driver categories. All models were trained on a single NVIDIA TITAN-Xp GPU.

\subsection*{Pre-Training Details}
We investigate the effect of randomly initializing the weights versus initializing the weights from models trained on a large land cover dataset. We pair Indonesia land cover categories manually classified by the Ministry of Environment Forestry in 2017 \cite{margono2016indonesia,moef} with Landsat 8 images created using the same procedure to minimize cloud cover described in the Satellite Imagery section. We include all tiles in Indonesia except for any tiles that overlapped with tiles in the driver dataset. The full dataset consists of 75,923 examples and is used to train the models for the pre-training experiments.

\end{document}